\documentclass[letterpaper]{article} 
\usepackage{aaai2026}  
\usepackage{times}  
\usepackage{helvet}  
\usepackage{courier}  
\usepackage[hyphens]{url}  
\usepackage{graphicx} 
\usepackage{natbib}  
\usepackage{caption} 

\usepackage[x11names]{xcolor} 

\usepackage{algorithm}
\usepackage{algorithmic}
\usepackage{multirow}
\usepackage{amsmath}
\usepackage{colortbl}
\usepackage{arydshln}
\usepackage{booktabs}
\usepackage{afterpage}
\usepackage{amssymb}
\usepackage{mathtools}
\usepackage{newfloat}
\usepackage{listings}
\DeclareCaptionStyle{ruled}{labelfont=normalfont,labelsep=colon,strut=off} 
\floatstyle{ruled}
\newfloat{listing}{tb}{lst}{}
\floatname{listing}{Listing}
\title{FANoise: Singular Value-Adaptive Noise Modulation for Robust Multimodal Representation Learning}
\author{
    Jiaoyang Li\equalcontrib,
    Jun Fang\equalcontrib,
    Tianhao Gao,
    Xiaohui Zhang,
    Zhiyuan Liu,\\
    Chao Liu\thanks{Corresponding author},
    Pengzhang Liu,
    Qixia Jiang\\
}
\affiliations{
    JD, Retail, Beijing, China\\
    \{lijiaoyang7, fangjun8, gaotianhao1, zhangxiaohui40, liuzhiyuan8, liuchao397, liupengzhang, jiangqixia\}@jd.com
}

\begin{document}

\maketitle

\begin{abstract}
Representation learning is fundamental to modern machine learning, powering applications such as text retrieval and multimodal understanding. However, learning robust and generalizable representations remains challenging. While prior work has demonstrated that active noise injection, a form of data augmentation, can enhance encoding performance, most existing methods rely on heuristic or static noise, overlooking the dynamic nature of feature distributions during training. In this work, we systematically study the role of noise in representation learning from both gradient-based and feature distribution perspectives, using InfoNCE loss as a representative example. Focusing on multimodal representation learning, we propose \textbf{FANoise}, a novel feature-adaptive noise injection strategy. By leveraging the dynamics of contrastive learning, FANoise effectively mitigates the negative impacts of noise while preserving its benefits. Under this theoretically grounded framework, comprehensive experiments demonstrate that FANoise consistently improves overall performance on multimodal tasks across various base VLM models.
\end{abstract}

\section{Introduction}

Representation learning, which aims to capture meaningful and transferable features from raw data, has become a cornerstone of modern machine learning. It plays a pivotal role across diverse applications, from text retrieval (e.g., ~\citep{bge_embedding,li2023gte}) to multimodal understanding (e.g., ~\citep{radford2021clip,jia2021align,jiang2024vlm2vec,wei2024uniir,ren2024vista,zhang2024magiclens,liu2022universal}). Despite its remarkable successes, learning robust and generalizable representations remains challenging.

In multimodal representation learning, contrastive learning frameworks such as CLIP~\citep{radford2021clip}, ALIGN~\citep{jia2021align}, SigLIP~\citep{zhai2023sigmoid}, and BLIP~\citep{li2022blip} have achieved significant progress. Most current multimodal embedding methods primarily focus on architectural innovations~\citep{li2022blip,wei2024uniir,ren2024vista,zhang2024magiclens,liu2022universal} or on enriching training data via data augmentation~\citep{zhang2024gme,chen2025mme5,zhou2024megapairs}, such as applying transformations to existing samples or generating new synthetic data. While data augmentation serves to increase data diversity and thereby improve model robustness, it does so by implicitly introducing variations or perturbations at the data level. In contrast, explicit noise injection strategies operate directly at the representation or feature level, offering a more controllable and theoretically analyzable approach to improving robustness and generalization. However, the systematic study of such representation-level noise injection, especially in complex multimodal settings, remains largely unexplored.

Recently, active noise injection approaches such as CLAE~\citep{ho2020contrastive}, SimCSE~\citep{gao2021simcse}, and NEFTune~\citep{jain2023neftune} have demonstrated the effectiveness of controlled noise injection strategies including adversarial noise, dropout masks, and feature perturbations in improving representation quality, particularly in unimodal scenarios. These studies highlight the crucial role of both explicit and implicit noise injection mechanisms in representation learning. However, most existing methods rely on heuristic or static noise augmentation schemes, without explicitly modeling the underlying feature distributions or adapting to the dynamic nature of training. This leads to several fundamental open questions:
\textit{
\begin{itemize}
\item What are the underlying mechanisms by which noise injection affects representation learning?
\item How can we design and adapt optimal noise injection strategies for different feature structures and learning scenarios?
\item How do theoretical principles of noise injection lead to practical improvements in robustness and generalization?
\end{itemize}
}
Addressing these questions is essential for developing a systematic and theoretically grounded framework for multimodal representation learning.

\begin{figure*}
    \centering
    \includegraphics[width=0.8\linewidth]{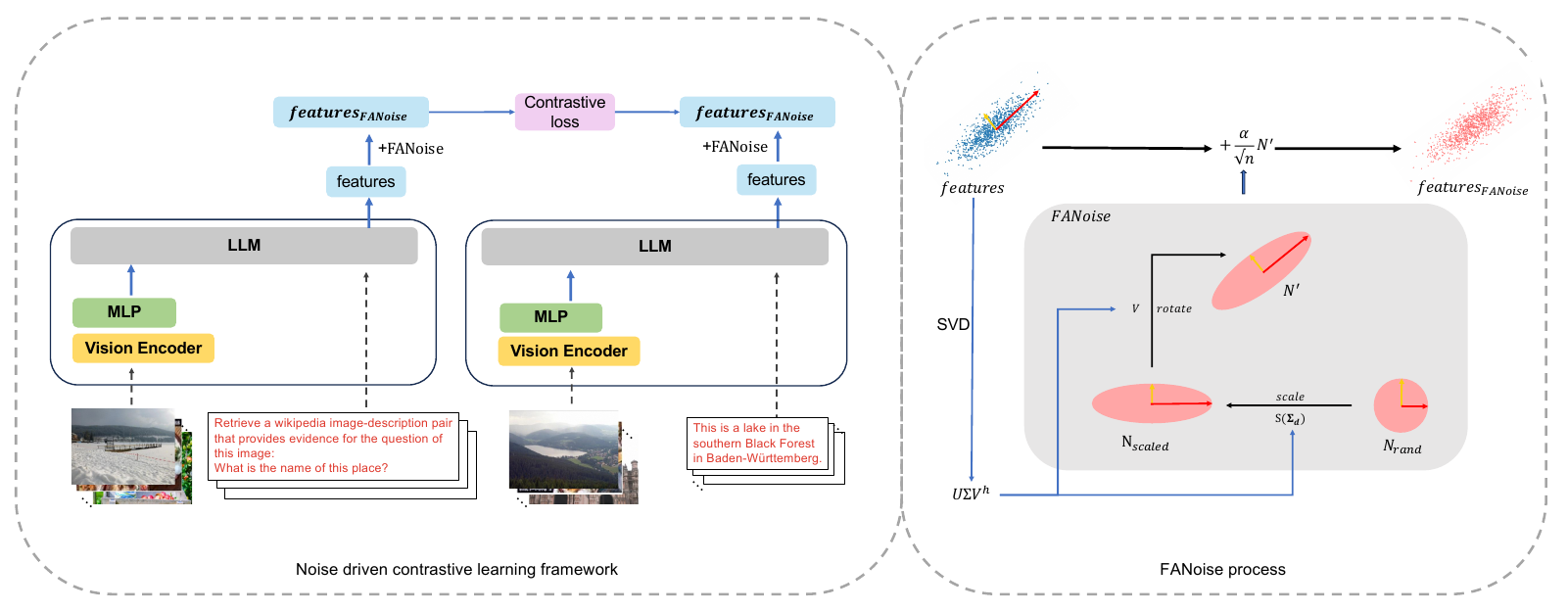}
    \caption{ A brief description of our idea. Left: we use a VLM as the backbone to deeply integrate image and text features, both query and target can be image, text or composed image-text data. Before the contrastive objective, we actively inject FANoise into features to improve representation quality. Right: detail of FANoise process, modulate Gaussian noise in V space by singular values after SVD and add it to the original features.}
    \label{fig:main-fig}
\end{figure*}

Motivated by these open questions and the limitations of existing approaches, this work systematically investigates the role of noise in multimodal representation learning from both gradient-based and feature distribution perspectives. Specifically, we analyze how noise injection affects the gradient dynamics of the InfoNCE loss and examine its impact on feature distributions through the lens of singular value decomposition (SVD). Based on these theoretical insights, we propose a novel \textbf{F}eature-\textbf{A}daptive \textbf{Noise} injection strategy, termed \textbf{FANoise}, which actively injects noise into features before the contrastive objective to improve representation quality. We provide an overview of FANoise in Fig.~\ref{fig:main-fig}.

Our contributions are threefold:
\begin{enumerate}
\item We provide a thorough empirical and theoretical analysis of how noise injection influences feature alignment and uniformity in multimodal contrastive learning, bridging the gap between empirical practice and theoretical understanding.
\item We systematically characterize the optimal noise strength and distribution for effective multimodal representation learning, revealing insights that previous heuristic approaches have overlooked.
\item We propose and validate a feature-adaptive noise augmentation strategy (FANoise) based on singular value decomposition, achieving significant improvements in downstream multimodal tasks.
\end{enumerate}

Through extensive experiments on established multimodal benchmarks, we demonstrate that our approach leads to more robust and generalizable representations, significantly outperforming existing methods. By clarifying the precise role of noise injection in multimodal representation learning, this work provides actionable guidelines for future research and practice.

\section{Related Works}

\paragraph{Multimodal Embedding} Early research employs dual-encoder models to extract and align features through contrastive loss, including CLIP~\citep{radford2021clip}, ALIGN~\citep{jia2021align}, SigLIP~\citep{zhai2023sigmoid}, and BLIP~\citep{li2022blip}. Later work explores feature integration strategies: UniIR~\citep{wei2024uniir} uses score fusion for separate embeddings, while VISTA~\citep{ren2024vista} enhances text encoders with visual fusion modules~\citep{zhang2024magiclens,liu2022universal}. However, these models struggle with deep-level integration, limiting performance in complex retrieval tasks~\citep{zhang2024gme}. 

Vision-Language Models (VLMs)~\citep{wang2024qwen2,abdin2024phi,liu2024llavanext} have emerged as popular backbones for multimodal embedding due to their ability to handle diverse image-text combinations and integrate features deeply within transformer architectures. E5-V~\citep{jiang2024e5v} and VLM2Vec~\citep{jiang2024vlm2vec} utilize instruction tuning to transform VLMs into embedding models. LLaVE~\citep{lan2025llave} enhances embeddings through hardness-weighted contrastive learning with reward models, while UniME~\citep{gu2025breaking} employs a two-stage framework with textual knowledge distillation and hard-negative enhanced tuning.

\paragraph{Contrastive Learning} Contrastive learning has achieved remarkable success in modern deep learning through advances in neural architectures and data augmentation techniques. Early unsupervised methods relied on generative approaches like autoencoders~\citep{vincent2008extracting}, which struggled with high-dimensional data. Instance discrimination~\citep{wu2018unsupervised} and contrastive objectives (e.g., CPC~\citep{oord2018representation}) demonstrated that pulling positive pairs together while pushing negatives apart yields transferable features without labels. Key innovations include memory banks~\citep{wu2018unsupervised}, momentum encoders~\citep{he2020momentum}, and temperature-scaled losses~\citep{chen2020simclr}. Subsequent work explores negative-free methods~\citep{grill2020bootstrap} and hybrid approaches combining contrastive learning with clustering~\citep{caron2020unsupervised} or masked modeling~\citep{bao2021beit}, extending beyond vision to NLP~\citep{logeswaran2018efficient}, graphs~\citep{velivckovic2018deep}, and multimodal tasks~\citep{radford2021clip}.

\paragraph{Noise Learning} Noise presents both challenges and opportunities in machine learning. Label noise with incorrect annotations is addressed through robust loss functions~\citep{2021zhangCELoss,wang2019symmetric,2023Dynamics}, regularization~\citep{shorten2019survey,szegedy2016rethinking,liu2020early}, label correction~\citep{lu2022selc,gong2022class,zhang2021learning}, and sample selection~\citep{kim2019nlnl,han2018co,xia2023combating}. Conversely, structured noise in embeddings enhances generalization when controlled. NEFTune~\citep{jain2023neftune} shows that injecting controlled noise into embeddings during fine-tuning improves LLM robustness. Similarly, stochastic weight averaging~\citep{izmailov2018swa} and manifold mixup~\citep{verma2019manifold} leverage noise to smooth decision boundaries. 

In generative modeling, Denoising Diffusion Probabilistic Models~\citep{ho2020denoising} systematically corrupt and reconstruct data, transforming noise into structured samples. For self-supervised learning, noise generates positive pairs for contrastive objectives: SimCSE~\citep{gao2021simcse} employs dropout noise for augmented views, while MoCo~\citep{he2020momentum} and BYOL~\citep{grill2020bootstrap} rely on noise-induced augmentations for invariant representations. Thus, noise is detrimental in supervised settings with label corruption but beneficial in embeddings, contrastive learning, and generative modeling.

\section{Method}
In this section, we conduct an in-depth analysis of the noise mechanism and propose a method to effectively incorporate noise with contrastive learning. 

\subsection{Noise Driven Contrastive Learning}

\paragraph{InfoNCE Loss}

InfoNCE Loss~\citep{oord2018infonce} has emerged as the de facto standard in representation learning, serving as a widely used loss function for contrastive learning. We use the standard InfoNCE loss as our training objective:
\begin{equation}
    L(\mathbf{q},\mathbf{k}) = -\sum_{l=1}^N\log\frac{\exp(\mathbf{q_l}\cdot \mathbf{k_l}/\tau)}{\sum_{i=1}^N{\exp(\mathbf{q_l}\cdot \mathbf{k_i}/\tau)}}.
\label{infonce_loss}
\end{equation}

Formally, for a batch of training data \{($x_1$,$y_1$),...,($x_N$,$y_N$)\}, the vectors $\mathbf{q_l}$ and $\mathbf{k_l}$ are the normalized embeddings of the $x_l$ and $y_l$, $\tau$ is temperature parameter. In this work, both $x_l$ and $y_l$ can be image, text, or composed image-text data.

\paragraph{The Gradient Of InfoNCE Loss}

To analyze the influence of gradients on representation, we derive the gradient of InfoNCE loss with respect to the model parameters $\boldsymbol{\theta}$:

\begin{equation}
    \nabla_{\boldsymbol{\theta}}{L(\mathbf{q},\mathbf{k})} = \sum_l^N\nabla_{\mathbf{q_l}} {L(\mathbf{q},\mathbf{k})} \nabla_{\boldsymbol{\theta}}{\mathbf{q_l}} + \sum_i^N\nabla_{\mathbf{k_i}} {L(\mathbf{q},\mathbf{k}) }\nabla_{\boldsymbol{\theta}}{\mathbf{k_i}} .
\end{equation}

Separately, we derive $\nabla_{\mathbf{q_l}}{L(\mathbf{q},\mathbf{k})}$ and $\nabla_{\mathbf{k_i}} {L(\mathbf{q},\mathbf{k})}$, for the detailed derivation in the Appendices:

\begin{subequations}
\begin{align}
    \nabla_{\mathbf{q_l}}{L(\mathbf{q},\mathbf{k})} &= -\mathbf{k_l}/\tau + \overline{\mathbf{k_l}}/\tau  \label{q_grad}, \\
    \nabla_{\mathbf{k_i}} {L(\mathbf{q},\mathbf{k})} &= -\mathbf{q_i}/\tau + \widetilde{\mathbf{q_i}}/\tau \label{k_grad}.
\end{align}
\end{subequations}

Here defines $\overline{\mathbf{k_l}} = \sum_{j=0}^N{p_{lj} \mathbf{k_j}}$, $\widetilde{\mathbf{q_j}} =  \sum_{l=0}^N{p_{lj} \mathbf{q_l}}$ and $p_{lj} = \exp(\mathbf{q_l}\cdot \mathbf{k_j}/\tau)/\sum_{i=0}^N{\exp(\mathbf{q_l}\cdot \mathbf{k_i}/\tau)}$. 
Notably, the normalization condition $\sum_{j=0}^N p_{lj} = 1$ holds for each $l$, while no such guarantee holds when summing over  $l$ for fixed $j$.  

Finally,  we get:

\begin{equation}
\begin{aligned}
    &\nabla_{\boldsymbol{\theta}}{L(\mathbf{q},\mathbf{k})} \\
    &=-\frac{1}{\tau}\left [ \sum_l^N{ (\mathbf{k_l} - \overline{\mathbf{k_l}})\nabla_{\boldsymbol{\theta}}{\mathbf{q_l}}} + \sum_i^N(\mathbf{q_i} - \widetilde{\mathbf{q_i}})\nabla_{\boldsymbol{\theta}}{\mathbf{k_i}} \right ].
\label{infonce_grad}
\end{aligned}
\end{equation}

The gradient's purpose here is evident - it reduces the distance between $\mathbf{q_i}$ and $\mathbf{k_i}$ while maximizing deviation from the overall weighted mean direction, thus enhancing the isotropy of the representation space.

\paragraph{Noise Driven InfoNCE Loss}

To demonstrate the function of slight noise, which causes fluctuations in the encoding of $\mathbf{q},\mathbf{k}$, we add 
Gaussian noise to $\mathbf{k}$ as an example for analysis and assume that the noise to $\mathbf{k_l}$ is $\boldsymbol{\epsilon_l}$. Based on Eq.(\ref{q_grad}), the derivative of InfoNCE Loss with respect to the noise is:

\begin{equation}
\begin{aligned}
    \nabla_{\mathbf{q_l}} {L(\mathbf{q},\mathbf{k}+\boldsymbol{\epsilon})} &=  -\frac{1}{\tau}(\mathbf{k_l} + \boldsymbol{\epsilon_l} - \overline{(\mathbf{k}+\boldsymbol{\epsilon})_l}).
\label{noised_q_grad}
\end{aligned}
\end{equation}

Based on Taylor expansion and expectation of noise distribution, Eq.(\ref{noised_q_grad}) can be expanded as:

\begin{equation}
\begin{aligned}
    \mathbf{E}_{\boldsymbol{\epsilon}\sim\mathcal{N}(0,\delta^2)}\left [\nabla_{\mathbf{q_l}}{L(\mathbf{q},\mathbf{k}+\boldsymbol{\epsilon})}\right ] &= -\frac{1}{\tau}(\mathbf{k_l} -\overline{\mathbf{k_l}} - \delta^2 \mathbf{q_l}/\tau),
\end{aligned}
\end{equation}

When the $\mathbf{k}$ sampling distribution is uniform enough, $\mathbf{k_l}$ is parallel to $\mathbf{q_l}$, assuming $\overline{\mathbf{k_l}} \approx \gamma_l \mathbf{q_l}$:

\begin{equation}
\label{formula:noise_accelerate}
\begin{aligned}
    \mathbf{E}_{\boldsymbol{\epsilon}\sim\mathcal{N}(0,\delta^2)}\left [\nabla_{\mathbf{q_l}}{L(\mathbf{q},\mathbf{k}+\boldsymbol{\epsilon})} \right ] &\approx -\frac{1}{\tau}[\mathbf{k_l} - (1+ \delta^2/\tau \gamma_l)\overline{\mathbf{k_l}} ]. \\
\end{aligned}
\end{equation}

Under the premise of InfoNCE loss, the noise term is equivalent to giving negative samples a larger weight. It also makes $\mathbf{q_l}$ moves away from $\overline{\mathbf{k}}$ and closer to $\mathbf{k_l}$ more quickly. 

Significantly, noise on $\mathbf{k}$ contributes no more terms in Eq.(\ref{k_grad}) under expectation of noise distribution:

\begin{equation}
\label{formula:noise}
\begin{aligned}
    \mathbf{E}_{\boldsymbol{\epsilon}\sim\mathcal{N}(0,\delta^2)}\left [\nabla_{\mathbf{k_i}}{L(\mathbf{q},\mathbf{k}+\boldsymbol{\epsilon})} \right ] = -\frac{1}{\tau}(\mathbf{q_i} - \widetilde{\mathbf{q_i}} ). \\
\end{aligned}
\end{equation}

The detailed derivation process for this section is provided in the Appendices.

\subsection{Feature-Adaptive Noise Distributions}

\paragraph{Motivation and Problem Statement}
Data augmentation through noise injection has proven essential for improving model robustness and generalization. However, conventional approaches apply uniform noise across all feature dimensions, ignoring the inherent heterogeneity in feature importance and signal strength. This uniform treatment leads to two critical issues: (1) the noise energy will increase linearly with the feature dimension, creating instability in high-dimensional spaces, and (2) important features with weak signal strength may be overwhelmed by noise, degrading their signal-to-noise ratio below the discrimination threshold and causing the loss of critical discriminative information. To address these challenges, we propose \textbf{FANoise} (Feature-Adaptive Noise injection), which adapts noise intensity according to the spectral characteristics of the data.

\paragraph{Spectral Structure Analysis}
Understanding the spectral properties of data is fundamental to our approach. Let $\mathbf{X} \in \mathbb{R}^{m \times n}$ denote a data matrix with $m$ samples and $n$ features. Through singular value decomposition (SVD), we obtain its spectral representation:
\begin{equation}
    \mathbf{X} = \mathbf{U} \Sigma \mathbf{V}^\top,
\end{equation}
where $\mathbf{U} \in \mathbb{R}^{m \times m}$ and $\mathbf{V} \in \mathbb{R}^{n \times n}$ are unitary matrices, and $\Sigma \in \mathbb{R}^{m \times n}$ contains non-negative singular values $\sigma_1 \geq \sigma_2 \geq \cdots \geq \sigma_r > 0$ ($r = \text{rank}(\mathbf{X})$). The columns of $\mathbf{V}$ define principal directions that capture feature correlations, while singular values quantify the energy concentration in each direction. This decomposition reveals the natural hierarchy of feature importance, with larger singular values corresponding to more significant patterns in the data.

\paragraph{Limitations of Conventional Noise Injection}
\label{dim_collapse}
Traditional data perturbation methods apply isotropic Gaussian noise uniformly across all dimensions:
\begin{equation}
    \mathbf{\widetilde{X}}_{\text{naive}} = \mathbf{X} + \alpha \mathbf{N},
    \label{naive_noise}
\end{equation}
where $\mathbf{N} \sim \mathcal{N}^{m\times n}(0, \mathbf{I})$ and $\alpha > 0$ controls noise magnitude. This approach suffers from a fundamental dimensional scaling problem. The noise covariance structure $\text{Cov}(\alpha \mathbf{N}) = \alpha^2 \mathbf{I}_n$ results in total noise energy $\text{tr}(\alpha^2 \mathbf{I}_n) = \alpha^2 n$, which grows linearly with feature dimension $n$. In high-dimensional spaces, this cumulative noise energy eventually overwhelms the signal, creating a dimensional curse that degrades model performance.

To mitigate this dimensional dependency, a common solution scales the noise injection as:
\begin{equation}
    \mathbf{\widetilde{X}} = \mathbf{X} + \frac{\alpha}{\sqrt{n}} \mathbf{N}.
    \label{normalized_noise}
\end{equation}
The $\frac{1}{\sqrt{n}}$ scaling factor ensures constant total noise energy $\alpha^2$, independent of dimension $n$. However, this uniform scaling still fails to account for the heterogeneous importance of different feature directions.

\paragraph{Theoretical Foundation for Adaptive Scaling}
The need for feature-adaptive noise injection becomes evident when analyzing how noise affects different spectral components. As noise strength increases, dimensions associated with smaller singular values suffer disproportionate signal-to-noise ratio (SNR) degradation. This phenomenon arises from three interconnected theoretical principles:

\textbf{Spectral Perturbation Theory:} According to Weyl's perturbation theorem \citep{tao2012topics}, for spectral norm-bounded noise $\mathbf{N}$, the singular values satisfy:
\begin{equation}
    \label{weyl}
    |\sigma_i(\mathbf{\widetilde{X}}) - \sigma_i(\mathbf{X})| \leq \sigma_0\left(\frac{\alpha}{\sqrt{n}}\mathbf{N}\right).
\end{equation}
This bound reveals that dominant features with $\sigma_i(\mathbf{X}) \gg \sigma_0(\frac{\alpha}{\sqrt{n}}\mathbf{N})$ remain stable under noise perturbation, while marginal features with $\sigma_i(\mathbf{X}) \approx \sigma_0(\frac{\alpha}{\sqrt{n}}\mathbf{N})$ risk complete SNR collapse.

\textbf{Random Matrix Theory:} The effective covariance structure of noisy data exhibits a phase transition phenomenon described by spiked covariance models \citep{perry2016optimal,perry2018optimality}:
\begin{equation}
    \label{spiked}
    \mathbf{\widetilde{X}}^\top\mathbf{\widetilde{X}} = \sum_{i=1}^r \sigma_i^2 \mathbf{v}_i\mathbf{v}_i^\top + \frac{m}{n} \alpha^2\mathbf{I}.
\end{equation}
This decomposition shows that singular values above the critical threshold $\tau^* = \sqrt{\frac{m}{n}}\alpha$ remain separable from the noise bulk described by the Marchenko-Pastur distribution \citep{marchenko1967distribution}, while those below become statistically indistinguishable from noise.

\textbf{Information-Theoretic Perspective:} From an information preservation standpoint, uniform noise injection treats all feature directions equally, potentially destroying valuable information in weaker but discriminative dimensions. This motivates the need for adaptive scaling that preserves the relative importance hierarchy while maintaining robustness benefits.

\paragraph{Feature-Adaptive Noise Injection Method}
Building on the theoretical foundation, we propose FANoise, which adapts noise intensity according to local signal strength in each principal direction. The method operates through a two-stage process that systematically addresses the feature importance preservation and dimensional scaling problem:

\textbf{Stage 1 - Spectral-Aware Noise Generation and Transformation:} We first generate noise in the principal component space with adaptive scaling based on local signal strength:
\begin{equation}
\label{formula:scaled}
\mathbf{N_{scaled}} = \mathbf{N_{rand}} \odot \mathbf{S}(\Sigma_d),
\end{equation}
where $\mathbf{N_{rand}} \sim \mathcal{N}^{m\times r}(0,1)$, $\Sigma_d \in \mathbb{R}^{r}$ contains the diagonal singular values, and $\mathbf{S}(\Sigma_d)$ is the adaptive scaling function that modulates noise intensity based on local signal strength. This adaptive scaling addresses the feature importance preservation problem by ensuring that noise energy is distributed according to the spectral structure.

The scaled noise is then transformed back to the original feature space:
\begin{equation}
\mathbf{N'} = \mathbf{N_{scaled}} \mathbf{V}^{\top},
\end{equation}
where $\mathbf{V}^{\top} \in \mathbb{R}^{r\times n}$ contains only the effective principal components, ensuring $\mathbf{N'} \in \mathbb{R}^{m\times n}$.

\paragraph{Scaling Function Design}
We investigate three principled scaling strategies, each with distinct theoretical motivations:

\begin{itemize}
\item \textbf{Uniform scaling}: $\mathbf{S} = \mathbf{1}$ (equivalent to Eq.(\ref{naive_noise}))

This serves as our baseline, applying equal noise intensity across all directions.

\item \textbf{Linear scaling}: $\mathbf{S} = \Sigma_{d}/\overline{\Sigma_{d}}$

This approach scales noise proportionally to signal strength, maintaining constant signal-to-noise ratios and preserving weaker features' discriminative capacity.

\item \textbf{Sublinear scaling}: $\mathbf{S} = \sqrt{\Sigma_{d}} / \overline{\sqrt{\Sigma_{d}}}$

This strategy provides a compromise between uniform and linear scaling, offering moderate protection for weaker features while ensuring sufficient perturbation for robustness.
\end{itemize}

where $\overline{\cdot}$ denotes same as Eq.(\ref{q_grad}).

\textbf{Stage 2 - Dimensionally-Normalized Noise Injection:} The final noise injection applies proper dimensional normalization:
\begin{equation}
\widetilde{\mathbf{X}} = \mathbf{X} + \frac{\alpha}{\sqrt{n}} \mathbf{N'},
\end{equation}
As discussed above, the $\frac{\alpha}{\sqrt{n}}$ ensures that the total noise energy remains constant regardless of the feature dimension $n$.

\section{Experiments}
\label{Experiments}

This section evaluates FANoise by detailing the experimental setup, comparing it with state-of-the-art baselines, and analyzing noise strength and distribution. Experiments on noise position patterns are deferred to the Appendices.

\subsection{Dataset Overview and Experiment Setting}

\paragraph{Dataset Overview} The MMEB \citep{jiang2024vlm2vec} \footnote{https://huggingface.co/datasets/TIGER-Lab/MMEB-train} \footnote{https://huggingface.co/datasets/TIGER-Lab/MMEB-eval} (Massive Multimodal Embedding Benchmark) is a comprehensive benchmark that comprises 36 datasets divided into four meta-tasks: classification, visual question answering, retrieval, and visual grounding. It has 20 in-distribution datasets for training and 36 for testing, with 20 being in-distribution (IND) and 16 out-of-distribution (OOD). We calculate Precision@1 of each dataset, which measures the proportion of top-ranked candidates that are positive samples. MMEB aims to provide a diverse framework for training and evaluating instruction following multimodal embedding models, assessing their generalization across tasks and domains.

\paragraph{Experiment Setting} Our implementation builds on the Qwen2-VL-2B \citep{wang2024qwen2} following the VLM2Vec \citep{jiang2024vlm2vec}. The adaptation pipeline incorporates LoRA \citep{hu2022lora} adapters (rank=8) for parameter-efficient fine-tuning, processing high-resolution images, each equivalent to 4096 token sequences, with a maximum of 100{,}000 samples per train dataset. Optimization uses a linear learning rate scheduler (initial rate=2e-5) over 2,000 training steps with 200 warmup iterations, executing on 256 samples per GPU batch. To address memory constraints, GradCache \citep{gao2021GradCache} is implemented with chunk sizes of 4 for query and candidate embeddings. Critical design choices regarding noise injection parameters are systematically analyzed in Section Result Analysis. Extended implementation details for alternative architectures are documented in Appendices.

\paragraph{Baselines} Following VLM2Vec, we experiment with different backbone models (Qwen2-VL-2B~\citep{wang2024qwen2}, LLaVA-NeXT~\citep{liu2024llavanext} and Phi3.5-V-4B~\citep{abdin2024phi}). Additionally, we benchmark our results against contemporary advanced models: CLIP~\citep{radford2021clip}, BLIP2~\citep{li2023blip}, UniIR~\citep{wei2024uniir}, MagicLens~\citep{zhang2024magiclens}, VLM2Vec~\citep{jiang2024vlm2vec}, UniME~\citep{gu2025breaking}, MegaPairs~\citep{zhou2024megapairs}, LLaVE~\citep{lan2025llave}. 

\subsection{Main Result}

\begin{table*}[t]

\tabcolsep=8pt
\begin{tabular}{lccccccc}
\toprule
\multirow{2}{*}{Model} & \multicolumn{4}{c}{Per Meta-Task Score} & \multicolumn{3}{c}{Avg Score} \\
\cmidrule(lr){2-5}
& Classification & VQA & Retrieval & Grounding & IND & OOD & Overall \\
\# Datasets & 10 & 10 & 12 & 4 & 20 & 16 & 36 \\
\midrule
\multicolumn{8}{c}{\textit{\textbf{Zero-shot on MMEB}}} \\
\hline
CLIP~ & 42.8 & 9.1 & 53.0 & 51.8 & 37.1 & 38.7 & 37.8 \\
BLIP2~ & 27.0 & 4.2 & 33.9 & 47.0 & 25.3 & 25.1 & 25.2 \\
UniIR(BLIP$_{FF}$)~ & 42.1 & 15.0 & 60.1 & 62.2 & 44.7 & 40.4 & 42.8 \\
UniIR(CLIP$_{SF}$)~ & 44.3 & 16.2 & 61.8 & 65.3 & 47.1 & 41.7 & 44.7 \\
Magiclens~ & 38.8 & 8.3 & 35.4 & 26.0 & 31.0 & 23.7 & 27.8 \\
\midrule
\multicolumn{8}{c}{\textit{\textbf{Fine-tuning on MMEB}}} \\
\hline
UniME(Phi3.5-V)~ & 54.6 & 55.9 & 64.5 & 81.8 & 68.2 & 52.7 & 61.3 \\
UniME(LLaVA-1.6)~ & 60.6 & 52.9 & 67.9 & 85.1 & 68.4 & 57.9 & 63.6 \\
MegaPairs(LLaVA-1.6)~& 56.0& 57.4& 69.9& 83.6& 68.0& 59.1&64.1\\
VLM2Vec(LLaVA-OV-7B)~& 63.5& 61.1& 64.5& 87.3& 69.7& 61.0&65.8\\
LLaVE(LLaVA-OV-7B)~& 65.7& 65.4& 70.9& 91.9& 75.0& 64.4&70.3\\

\hdashline 
VLM2Vec(Phi3.5-V)~ & 54.8 & 54.9 & 62.3 & 79.5 & 66.5 & 52.0 & 60.1 \\
FANoise$_{ss}$(Phi3.5-V) & 54.6 & \textbf{56.1} & \textbf{63.9} & 78.6 & \textbf{68.2} & 51.5 & \textbf{60.8(+0.7)}\\
VLM2Vec(Qwen2-VL-2B)~ & 59.0 & 49.4 & 65.4 & 73.4 & 66.0 & 52.6 & 60.1 \\
FANoise$_{ss}$(Qwen2-VL-2B) & \textbf{60.3} & \textbf{51.3} & \textbf{66.2} & 72.3 & \textbf{67.2} & \textbf{53.4} & \textbf{61.1(+1.0)} \\
VLM2Vec(LLaVA-1.6-LR)~ & 54.7 & 50.3 & 56.2 & 64.0 & 61.0 & 47.5 & 55.0 \\
FANoise$_{ss}$(LLaVA-1.6-LR) & \textbf{57.0} & 43.1 & \textbf{64.1} & \textbf{90.1} & \textbf{66.1} & \textbf{50.5} & \textbf{59.2(+4.2)} \\
VLM2Vec(LLaVA-1.6-HR)~ & 61.2 & 49.9 & 67.4 & 86.1 & 67.5 & 57.1 & 62.9 \\
FANoise$_{ss}$(LLaVA-1.6-HR) &  \textbf{62.2}&  \textbf{57.3}&  \textbf{69.1}&  \textbf{91.4}&  \textbf{72.2}&  \textbf{59.1}&  \textbf{66.4(+3.5)} \\
 VLM2Vec(Qwen2-VL-7B)~& 62.6& 57.8& 69.9& 81.7& 72.2& 57.8&65.8\\
FANoise$_{ss}$(Qwen2-VL-7B) &  \textbf{63.7}&  \textbf{59.4}&  \textbf{70.3}&  80.4&  \textbf{73.5}&  \textbf{57.9}&  \textbf{66.6(+0.8)} \\
\bottomrule
\end{tabular}
\caption{
Performance comparison on the MMEB benchmark.
The FF and SF subscripts under CLIP or BLIP represent feature-level fusion and score-level fusion, respectively. 
LR suffixes signify training and inference on low-resolution (336$\times$336) images, while HR suffixes similarly denote both processes on high-resolution (1344$\times$1344) images.
Reported scores are the average Precision@1 over the corresponding datasets. 
}

\label{table:main-result}
\end{table*}

In Table~\ref{table:main-result}, we present the performance of our method, where FANoise\(_{ss}\) represents \textit{sublinear scaling} noise, achieving the best results among the three distributions as verified in Section Noise Distribution. Following the experimental setup of VLM2Vec~\citep{jiang2024vlm2vec}, \textbf{FANoise$_{ss}$} achieves an average improvement of 2.04\% across five backbones (Phi3.5-V, Qwen2-VL-2B~\footnote{https://huggingface.co/TIGER-Lab/VLM2Vec-Qwen2VL-2B}, LLaVA-1.6-LR, LLaVA-1.6-HR and Qwen2-VL-7B), with notable improvements of $4.2\%$ for LLaVA-1.6-LR and $3.5\%$ for LLaVA-1.6-HR.

The best-performing model, FANoise$_{ss}$ (LLaVA-1.6-HR), outperforms contemporary SOTAs (UniME~\citep{gu2025breaking}, MegaPairs~\citep{zhou2024megapairs}, VLM2Vec~\citep{jiang2024vlm2vec}) by 0.6\% over the best-performing Qwen2-VL-7B~\footnote{https://huggingface.co/TIGER-Lab/VLM2Vec-Qwen2VL-7B}. Our UniME~\citep{gu2025breaking} results are from re-evaluating their checkpoint, as their original 'average overall score' wasn't a simple average across all 36 datasets.

LLaVA-OV-7B serves as a strong baseline, achieving comparable performance to Qwen2-VL-7B (65.8\%) in VLM2Vec~\citep{lan2025llave}. LLaVE~\citep{lan2025llave} further optimizes it to reach 70.3\%. Due to GPU constraints, we have not extended experiments to LLaVA-OV-7B~\citep{li2024onevision}.

Both LLaVE and UniME enhance discriminative power by prioritizing hard-negative samples during training; LLaVE uses a reward model for higher weights, while UniME filters false negatives and samples hard negatives. As shown by Eq.(\ref{formula:noise_accelerate}), FANoise effectively weights negative samples by noise, aligning with this approach. Additionally, it excels in simplicity, effectiveness, and integration ease:

\begin{enumerate}
    \item Generalized Robustness Across Architectures: FANoise enhances representation robustness and generalizes well to various model architectures and capacities.
    \item Seamless Integration: Its plug-and-play design integrates seamlessly into existing training pipelines without redesign costs.
\end{enumerate}

\subsection{Result Analysis}
\label{Result Analysis}

To enable flexible and efficient result analysis, we use a batch size of 16 per GPU on 8 GPUs with up to 5{,}000 training samples per dataset, employing Qwen2-VL-2B~\citep{wang2024qwen2} as the backbone model.

We identify an optimal range of noise strength that balances robustness with precision and show that our feature-adaptive noise consistently outperforms uniform Gaussian baselines. These results validate our critical design choices.

\subsubsection{Noise Strength}
\label{Noise Strength}

\begin{figure}[t]
    \centering
    \includegraphics[width=0.8\linewidth]{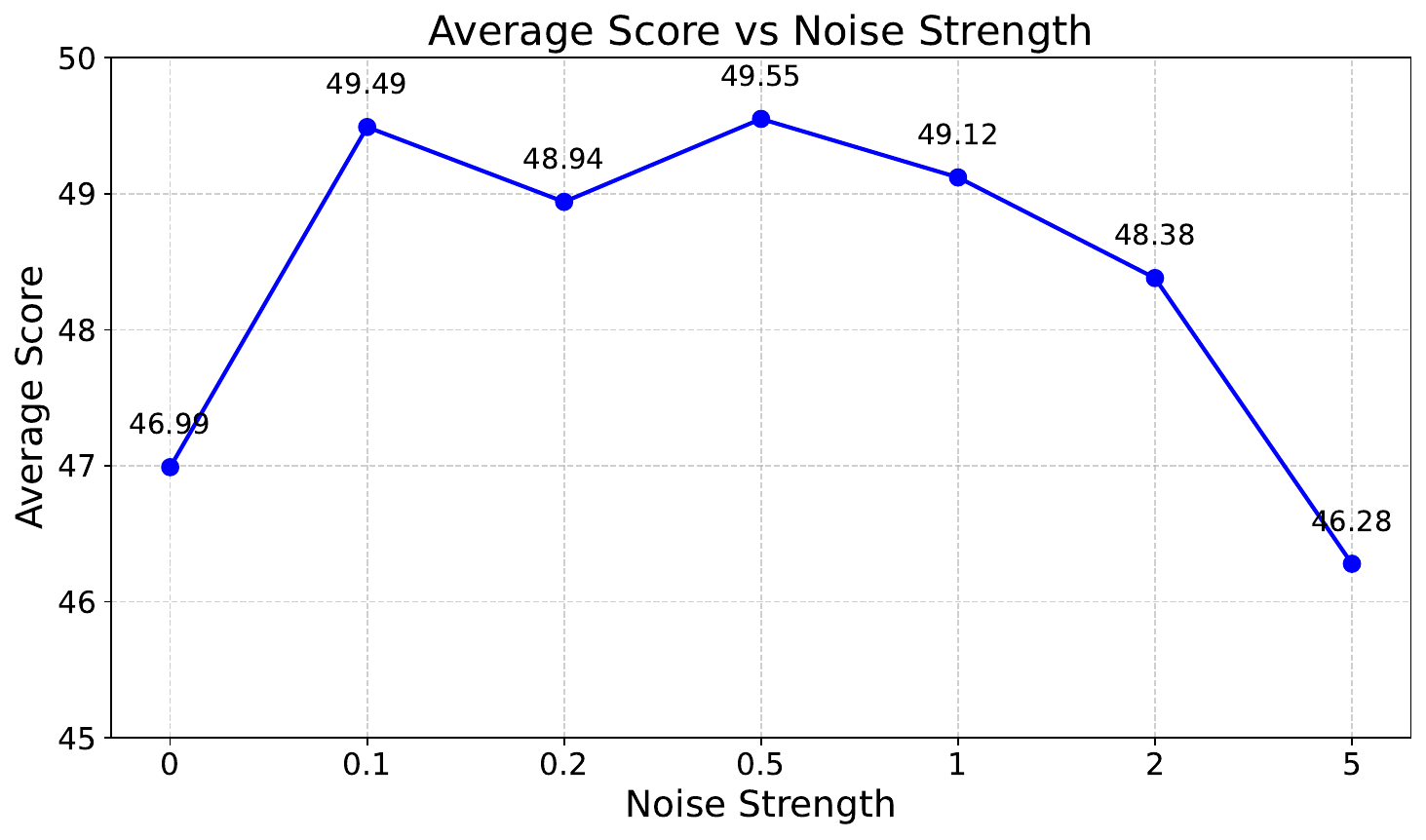}
    \caption{Performance comparison of different noise strengths on the MMEB benchmark}
    \label{fig:noise-strength}
\end{figure}

\begin{figure}[t]
    \centering
    \includegraphics[width=0.75\linewidth]{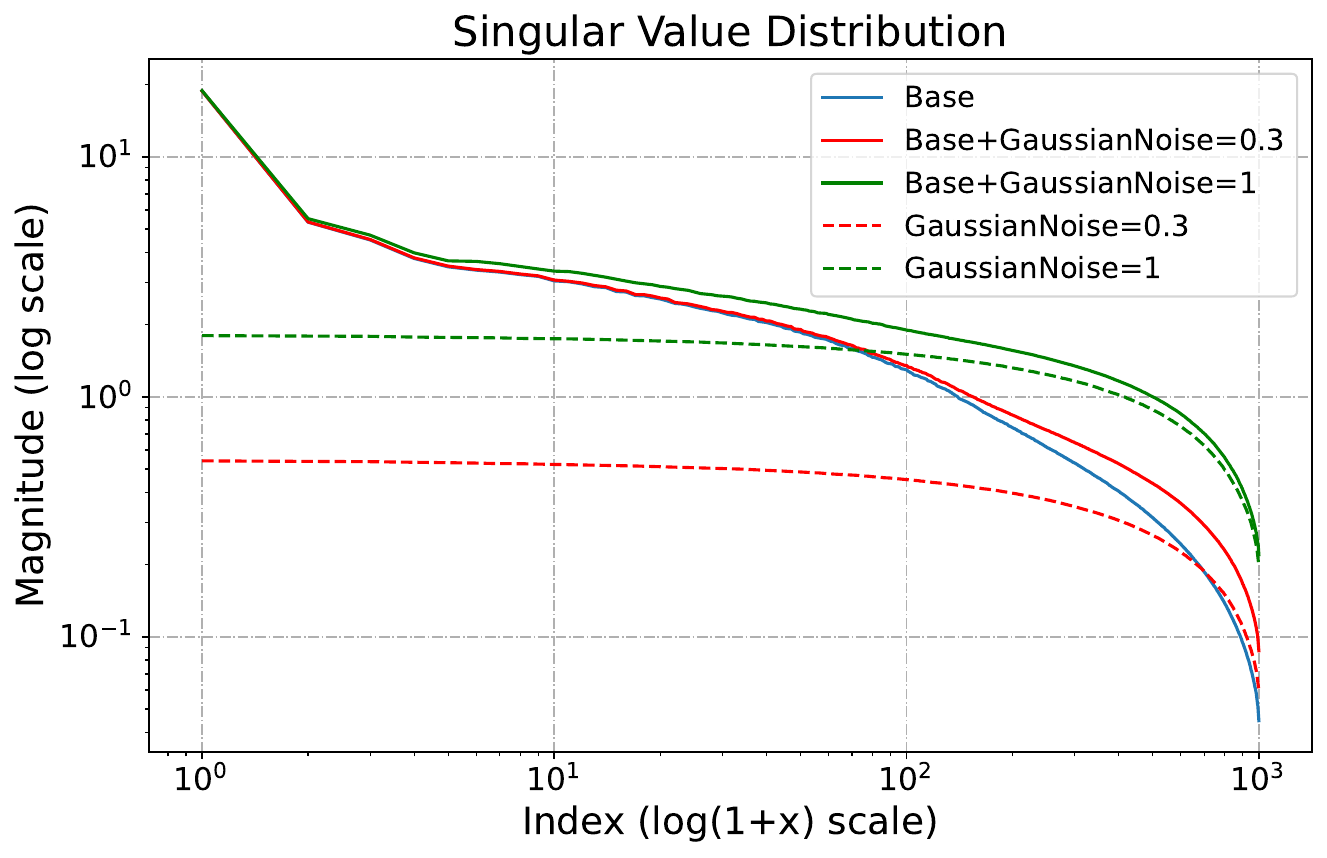}
    \caption{Log-log Plot of Singular Values: Base Embedding, Gaussian Noise, and Noisy Embedding}
    \label{fig:Base_GaussianNoise}
\end{figure}

Based on FANoise\(_{ss}\), we examine the impact of noise strength (\(\alpha\)) on model performance, as illustrated in Fig.~\ref{fig:noise-strength}. All experiments show improvements over the no-noise baseline except at overly high noise strength, indicating that excessive noise destroys tail features. Performance initially improves with increasing \(\alpha\) before declining.

To explore the interaction between noise strength and features, we disturb base embeddings with uniform Gaussian noise controlled by $\alpha/\sqrt{n}$, where $\alpha$ is set to \textbf{0.3} and \textbf{1}. Using 1000 randomly sampled test data points with feature dimension $n=1536$, we perform Singular Value Decomposition (SVD) on three matrices:

\begin{itemize}
    \item Base feature matrix($F$): Extracted from test data points;
    \item Pure Gaussian noise matrix($GN$): Same dimension as embeddings, with entries from $\mathcal{N}(0, (\alpha/\sqrt{n})^2)$;
    \item Noisy embedding matrix($F+GN$): Generated by adding scaled noise to base embeddings.
\end{itemize}

The singular values of these matrices are plotted on a log-log scale, with the x-axis representing the index of singular values and the y-axis their magnitudes.

\paragraph{Spectral Perturbation Analysis}
As observed in Fig.~\ref{fig:Base_GaussianNoise}, the singular values of $GN$ match the theoretical Marchenko-Pastur noise bulk, with boundary $[\left |1-\sqrt{\frac{m}{n}}\right |, 1+\sqrt{\frac{m}{n}}]*\alpha \approx [0.2, 1.8]*\alpha$. According to Eq.(\ref{weyl}), singular values of $F$ greater than the upper boundary ($1.8*\alpha$) effectively \textit{escape} the Marchenko-Pastur noise bulk, preserving their signal-to-noise ratio. When singular values fall below this boundary, $F$ experiences significant perturbations and enters a noise-dominated regime. This demonstrates how Weyl's perturbation theorem predicts the stability of dominant features under noise perturbation, while marginal features with comparable singular values to the noise spectral norm risk complete SNR collapse.

\begin{figure}
    \centering
    \includegraphics[width=0.7\linewidth]{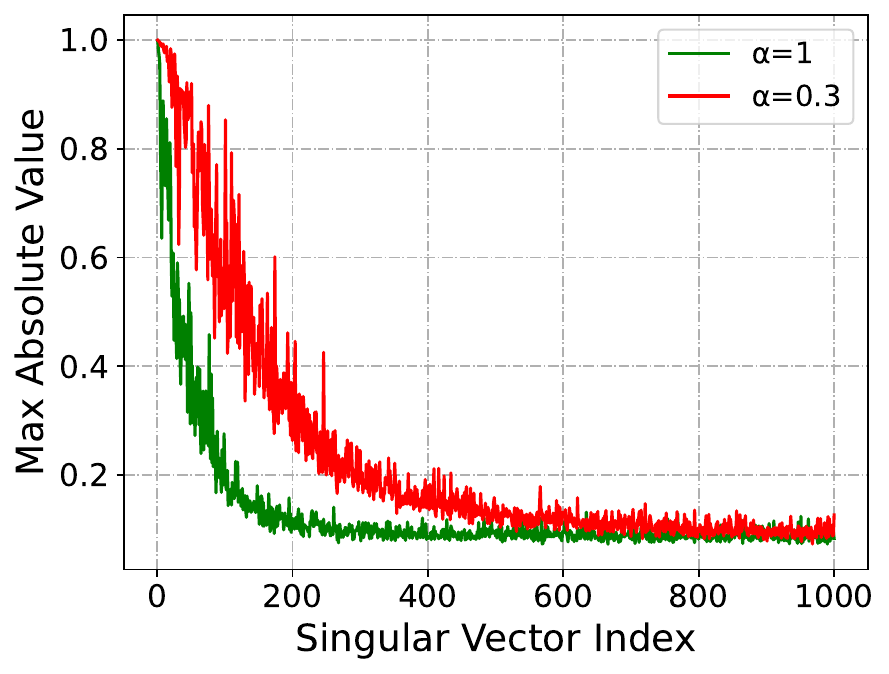}
    \caption{inner product between singular vector of $F$ and $F+GN$}
    \label{fig:base_add_noise_sim_alphais1_0p3}
\end{figure}

\paragraph{Phase Transition Analysis}
Additionally, Eq.(\ref{spiked}) provides a more precise critical threshold:
$\tau^* = \sqrt{\frac{m}{n}}\alpha \approx 0.8\alpha$.
This threshold marks the boundary between distinguishable (signal-dominated) and indistinguishable (noise-dominated) singular directions. For $\alpha=0.3/1$, the singular value around $0.24/0.8$ corresponds to index $600/180$. As shown in Fig.~\ref{fig:base_add_noise_sim_alphais1_0p3}, starting from the $600/180$-th singular value, the inner products between corresponding singular vectors of ($F$) and ($F + GN$) remain almost identical, indicating these singular directions become indistinguishable from noise.

These conclusions align closely with theoretical predictions, clearly elucidating how noise strength impacts the structural integrity of embedding features.
In summary, We select $\alpha = 0.1$ for our main experiments, as it consistently achieves competitive performance across various small-batch trials. This value represents a balance between regularization and information disruption, providing sufficient perturbation to prevent overfitting while preserving meaningful signal.

\subsubsection{Noise Distribution}
\label{Noise Distribution}

\begin{table}[t]
\centering
\begin{tabular}{cc}
    \hline
    Model &Avg Score \\
    \hline
    Baseline &60.06 \\
    Baseline + Uniform scaling &60.93(+0.87) \\
    Baseline + Linear scaling &60.69(+0.63) \\
    Baseline + Sublinear scaling &\textbf{61.08(+1.02)} \\
    \hline
\end{tabular}
\caption{Three Possible Noise Distribution Results on the MMEB benchmark}
\label{fig:big-trials}
\end{table}

\begin{figure}[t]
\centering
\includegraphics[width=\linewidth]{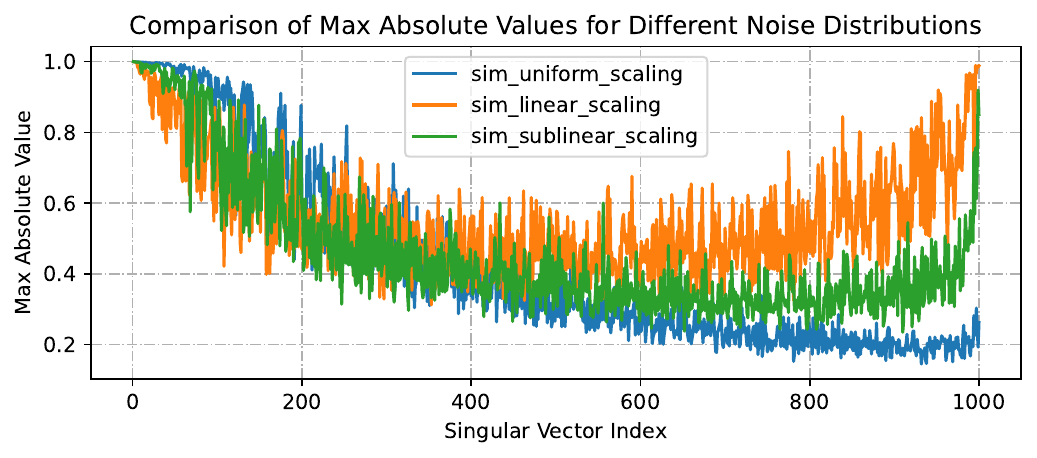}
\caption{Max Similarity Values: Base Embedding vs. Noisy Embeddings}
\label{fig:noisy-emb-sim-base}
\end{figure}

Due to the importance of noise distribution, we conduct full setup experiments on three noise distributions \(\mathbf{S}(\Sigma_d)\) of Eq.(\ref{formula:scaled}), all using fixed noise strength of 0.1. As shown in Fig.~\ref{fig:big-trials}, the baseline VLM2Vec~\footnote{https://huggingface.co/TIGER-Lab/VLM2Vec-Qwen2VL-2B} achieves 60.06\% on the MMEB benchmark. 
Uniform Gaussian noise (\textit{Uniform scaling}) improves performance by 0.87\% (60.93\%), suggesting that appropriate noise enhances model robustness by preventing overfitting. \textit{Linear scaling} shows minimal improvement, demonstrating that the absence of the square root step leads to suboptimal noise scaling. FANoise's \textit{Sublinear scaling} achieves the highest score of 61.08\%, outperforming the baseline by 1.02\%.
Under isotropic Gaussian noise (\textit{Uniform scaling}), dimensions with smaller singular values undergo disproportionately larger signal-to-noise ratio (SNR) drops due to weaker signal strength. FANoise protects discriminative features prone to SNR loss.

To analyze why \textit{sublinear scaling} performs best, we compute right-singular vectors $(\mathbf{V}^\top)$ via SVD for base and three noise-augmented embeddings, comparing their similarity to the base embedding (see Fig.~\ref{fig:noisy-emb-sim-base}). 
\textit{Uniform scaling} shows limited perturbation in higher singular value regions, while \textit{linear scaling} maintains high similarity across lower singular value regions. \textit{Sublinear scaling} demonstrates a different pattern, with moderate perturbation in higher singular value regions and reduced similarity in lower singular value regions, resulting in superior performance.

\section{Conclusion}
While most existing noise injection methods rely on heuristic or static approaches that fail to account for the dynamic nature of feature distributions during training, we conduct a systematic investigation from both gradient-based and feature distribution perspectives. Focusing on multimodal representation learning, we propose FANoise - a feature-adaptive noise injection strategy that dynamically responds to evolving feature structures. Our theoretical framework and extensive experiments demonstrate that FANoise consistently enhances robustness and generalization across models of varying architectures and capacities. Designed as a plug-and-play module, FANoise provides an architecture-agnostic solution that complements existing representation learning techniques, offering new possibilities for developing more adaptive and robust learning systems.


\appendix

\section{Appendices}

\subsection{Formula Derivation}
\label{Formula Derivation}
\subsubsection{The Gradient of InfoNCE Loss}
\label{The Gradient of InfoNCE Loss}
\begin{equation}
\begin{aligned}
    \nabla_\theta{L(q,k)} &= \sum_l^N\nabla_{q_l} {L(q,k)} \nabla_\theta{q_l} + \sum_i^N\nabla_{k_i} {L(q,k) }\nabla_\theta{k_i} 
\end{aligned}
\end{equation}
where $\nabla_{q_l} {L(q,k)}$ means the gradient of the InfoNCE loss with respect to the query vector $q_l$, $\nabla_\theta{q_l}$ is the gradient of the query vector $q_l$ with respect to model parameters $\theta$. $\sum_l^N\nabla_{q_l} {L(q,k)} \nabla_\theta{q_l}$ means the total query-side gradient contribution, summing over all N query representations, this term updates $\theta$ to improve the query embedding. Likewise, $\sum_i^N\nabla_{k_i} {L(q,k) }\nabla_\theta{k_i}$ adjusts $\theta$ to improve the key representations based on their alignment with queries.
\begin{equation}
\begin{aligned}
    \nabla_{q_l}{L(q,k)} &= -\nabla_{q_l}{\log\frac{\exp{(q_l\cdot k_l/\tau)}}{\sum_{i=0}^N{\exp{(q_l\cdot k_i/\tau)}}}} \\
    &= -\nabla_{q_l}{(q_l\cdot k_l/\tau)} + \nabla_{q_l}{(\log{\sum_{i=0}^N{\exp{(q_l\cdot k_i/\tau)}}})} \\
    &= -k_l/\tau + \sum_{j=0}^N\frac{\exp{(q_l\cdot k_j/\tau)}k_j/\tau}{\sum_{i=0}^N{\exp{(q_l\cdot k_i/\tau)}}} \\
    &= -k_l/\tau + \sum_{j=0}^N{p_{lj} k_j/\tau} \\
    &= -k_l/\tau + \overline{k_l}/\tau
\end{aligned}
\end{equation}

\begin{equation}
    \begin{aligned}
        \nabla_{k_i} {L(q,k)} &= -q_i/\tau + \sum_{l=0}^N{p_{lj} q_l/\tau} \\
        & = -q_i/\tau + \widetilde{q_i}/\tau
    \end{aligned}
\end{equation}

\begin{equation}
    \begin{aligned}
        p_{lj} &= \frac{exp(q_l\cdot k_j/\tau)}{\sum_{i=0}^N{exp(q_l\cdot k_i/\tau)}} \\
        \overline{k_l} &= \sum_{j=0}^N{p_{lj} k_j} \\
        \widetilde{q_j} &=  \sum_{l=0}^N{p_{lj} q_l}
    \end{aligned}
\end{equation}
Notably, the normalization condition $\sum_{j=0}^N p_{lj} = 1$ holds for each $l$, however, no such guarantee holds when summing over  $l$ for fixed $j$: $\sum_{l=0}^N{p_{lj}}\neq1$. Finally, we get:

\begin{equation}
\begin{aligned}
    \nabla_\theta{L(q,k)} &= \sum_l^N\nabla_{q_l} {L(q,k)} \nabla_\theta{q_l} + \sum_i^N\nabla_{k_i} {L(q,k) }\nabla_\theta{k_i} \\
    &=-\frac{1}{\tau}\left [ \sum_l^N{ (k_l - \overline{k_l})\nabla_\theta{q_l}} + \sum_i^N(q_i - \widetilde{q_i})\nabla_\theta{k_i} \right ]
\end{aligned}
\end{equation}

\subsubsection{The Gradient of InfoNCE Loss when Adding Noise to k}
\label{The Gradient of InfoNCE Loss when Adding Noise to k}

Assuming that noise to $k_l$ is $\epsilon_l$, the derivative of InfoNCE Loss with respect to the model parameters is:

\begin{equation}
\begin{aligned}
    \nabla_{q_l} {L(q,k)} &=  -\frac{1}{\tau}(k_l + \epsilon_l - \overline{(k+\epsilon)_l})
\end{aligned}
\end{equation}
Based on Taylor expansion and the expectation of noise distribution, \(\overline{(k+\epsilon)_l}\) can be expanded as Eq.(\ref{noise_to_g_drad_detail}). Note that in Eq.(\ref{noise_to_g_drad_detail}), the subscript \(\epsilon\sim\mathcal{N}(0,\delta^2)\) is omitted for typesetting clarity, though all expectations are taken with respect to this distribution.

\begin{equation}
\begin{aligned}
    \MoveEqLeft \mathbf{E}_{\epsilon\sim\mathcal{N}(0,\delta^2)}\left [\overline{(k+\epsilon)_l} \right ] \\
    &= \mathbf{E}_{\epsilon\sim\mathcal{N}(0,\delta^2)}\left [\sum_{i=0}^N{\frac{\exp(q_l\cdot (k_i+\epsilon_i)/\tau)}{\sum_{j=0}^N{\exp(q_l\cdot (k_j + \epsilon_j)/\tau)}}(k_i + \epsilon_i)} \right ]\\
    &\approx \mathbf{E}\left [\sum_{i=0}^N{\frac{[1+\epsilon_{i,q_l}/\tau + \epsilon_{i,q_l}^2/(2\tau^2)] \exp(q_l\cdot k_i/\tau)(k_i + \epsilon_i)}{\sum_{j=0}^N{[1+\epsilon_{j,q_l} / \tau + \epsilon_{j,q_l}^2/(2\tau^2)] \exp(q_l\cdot k_j /\tau)}}} \right ] \\
    &= \mathbf{E}\left [\sum_{i=0}^N{\frac{[1+\epsilon_{i,q_l}/\tau + \epsilon_{i,q_l}^2/(2\tau^2)] p_{li} (k_i + \epsilon_i)}{\sum_{j=0}^N{[1+\epsilon_{j,q_l} / \tau + \epsilon_{j,q_l}^2/(2\tau^2)]p_{lj}}}} \right ] \\
    &= \mathbf{E}\left [\sum_{i=0}^N{\frac{[1+\epsilon_{i,q_l}/\tau + \epsilon_{i,q_l}^2/(2\tau^2)] p_{li} (k_i + \epsilon_i)}{1 + \sum_{j=0}^N{[\epsilon_{j,q_l} / \tau + \epsilon_{j,q_l}^2/(2\tau^2)]p_{lj}}}} \right ] \\
    &\approx \mathbf{E}\left [\sum_{i=0}^N{[1+\epsilon_{i,q_l}/\tau + \epsilon_{i,q_l}^2/(2\tau^2)] [1 - \delta^2/(2\tau^2)] p_{li} (k_i + \epsilon_i)} \right ] \\
    &= \mathbf{E}\left [\sum_{i=0}^N{[1+\epsilon_{i,q_l}/\tau + \epsilon_{i,q_l}^2/(2\tau^2) - \delta^2/(2\tau^2) ] p_{li} (k_i + \epsilon_i)} \right ] \\
    &= \mathbf{E}\left [\sum_{i=0}^N{(1 +\epsilon_{i,q_l}/\tau) p_{li} (k_i + \epsilon_i)} \right ] \\
    &= \mathbf{E}\left [\sum_{i=0}^N{(p_{li} k_i + p_{li} \epsilon_{i,q_l}^2 q_l)} \right ] \\
    &= \overline{k_l} + \delta^2 q_l/\tau 
    \label{noise_to_g_drad_detail}
\end{aligned}
\end{equation}

where we decompose $\epsilon$ as $\epsilon = \epsilon_q q + \epsilon_{q_{\perp}} q_{\perp}$, and the isotropy of Gaussian distribution implies that:

\begin{equation}
    \begin{aligned}
        \mathbf{E}_{\epsilon\sim\mathcal{N}(0,\delta^2)}\left [ \epsilon_{i,q} \epsilon_{i,q_{\perp}} \right ] &= 0 \\
        \mathbf{E}_{\epsilon\sim\mathcal{N}(0,\delta^2)}\left [\epsilon_{i,q}^2 \right ] &= \delta^2
    \end{aligned}
\end{equation}

Similar to Eq.(\ref{noise_to_g_drad_detail}), $\widetilde{q}$ under $\epsilon$ can be wrote as :

\begin{equation}
\begin{aligned}
    \MoveEqLeft \mathbf{E}_{\epsilon\sim\mathcal{N}(0,\delta^2)}\left [\widetilde{q_i} \right ] \\
    &= \mathbf{E}_{\epsilon\sim\mathcal{N}(0,\delta^2)}\left [\sum_{l=0}^N{\frac{\exp(q_l\cdot (k_i+\epsilon_i)/\tau)}{\sum_{j=0}^N{\exp(q_l\cdot (k_j + \epsilon_j)/\tau)}}q_l } \right ] \\
    &= \widetilde{q_i}
\end{aligned}
\end{equation}

\subsection{Experiment on Noise Position}
\label{Noise Position}

\begin{table}[t]
    \centering
    \begin{tabular}{lcc}
        \toprule
        Model & Noise Position & Avg Score \\
        \midrule
        baseline & - & 46.99 \\
        uniform Gaussian & Input layer & 48.46(+1.47) \\
        uniform Gaussian & Output layer & 48.73(+1.75) \\
        \bottomrule
    \end{tabular}
    \caption{Noise position results on the MMEB benchmark}
    \label{table:noise-position}
\end{table}

To evaluate the impact of noise injection strategies, we compare adding uniform Gaussian noise to the input layer (embedding layer outputs) versus the output layer (the final token embeddings of VLM's last hidden state) with the same noise strength (0.1). As shown in Table~\ref{table:noise-position}, the noise in the output layer consistently outperforms the noise in the input layer. We attribute this superiority to the following factors:
\begin{itemize}
    \item Reduced Information Loss: Noise added to the input layer can be partially eliminated or weakened as it passes through multiple layers of the network. In contrast, noise closer to the output layer can directly influence the final feature representation without being diluted by intermediate layers;
    \item Direct Impact on Learning Objectives: Noise closer to the loss computation directly affects the similarity calculations between positive and negative pairs, thereby having a more immediate impact on the learning objective;
    \item Enhanced Feature Robustness: Adding noise near the output layer enhances robust feature learning, ensuring stable outputs despite minor perturbations, thus improving generalization and reliability.
\end{itemize}

\subsection{Additional Experimental Details}
\label{Additional Experimental Details}

\paragraph{FANoise Algorithm Details}
During training, the noise modulation process is shown in Algorithm 1. For each batch of data,  we feed query and target separately into VLM. The final hidden state of the last token from each is used as input of the noise modulation process. Features augmented by the FANoise serve as input for the InfoNCE loss.

\paragraph{Additional Training Details}
\label{Additional Training Details}
All our models are trained using 8\(\times\)H800  80G GPUs, with 800G of memory. We train FANoise models based on different backbone models with a random seed of 42, and evaluate without noise. In Table~\ref{tab:training hyperparameters}, we demonstrate the training configurations of the FANoise based on three backbone models: Qwen2-VL-2B~\footnote{https://huggingface.co/Qwen/Qwen2-VL-2B-Instruct}, Phi3.5-V-4B~\footnote{https://huggingface.co/microsoft/Phi-3.5-vision-instruct}, and LLaVA-1.6~\footnote{https://huggingface.co/llava-hf/llava-v1.6-mistral-7b-hf}.

\paragraph{Training Speed Analysis} In our experiments using LLaVA-1.6~\citep{liu2024llavanext} on low-resolution images, we conduct two training protocols: a baseline group without noise addition and our proposed FANoise-based approach. Each configuration was trained for 1{,}000 steps, requiring approximately 65 hours of compute time, indicating that the inclusion of FANoise incurs negligible overhead. We attribute this efficiency to the noise being introduced only at the output layer features, where the singular value decomposition (SVD) step introduces a minimal computational burden relative to the overall training process.

\paragraph{Specific results on the MMEB}
We present comparative results of eight models on the MMEB benchmark in Table~\ref{tab:all mmeb result}. The performance metrics for CLIP, BLIP-2, UniIR, and MagicLens are sourced directly from VLM2Vec~\citep{jiang2024vlm2vec}. Conversely, our reported results for UniME~\citep{gu2025breaking} are based on re-evaluations using their published checkpoint, as their original "average overall score" calculation did not employ a straightforward average across the 36 datasets.

\subsection{Limitations}
\label{Limitations}

Although this work has demonstrated the effectiveness of the proposed feature-adaptive noise injection strategy in improving robustness and generalization within multimodal representation learning on the MMEB benchmark, its application scope is currently limited to this specific domain. For other representation learning scenarios, such as text retrieval, further experiments are needed to validate the effectiveness of our approach. Future work will explore these additional domains to provide broader insights into enhancing the robustness of representation learning models.

\newpage
\begin{figure*}[t]  
  \centering
  \begin{minipage}{.95\textwidth}  
    \begin{algorithm}[H]  
      \caption{Feature-Adaptive Noise injection}
      \label{alg:full}
      \begin{algorithmic}[1]
    \REQUIRE Input embeddings matrix
               $\mathbf{E}\in\mathbb{R}^{B\times d}$ ($B$ batch size, $d$ embedding dimension);
               noise strength $\alpha\,({=}0.1)$
    \ENSURE  Noise-augmented embeddings
               $\mathbf{E}'\in\mathbb{R}^{B\times d}$.
    \STATE $\boldsymbol{\varepsilon}\sim\mathcal{N}(0,1)\in\mathbb{R}^{B\times d}$
           \hfill // sample i.i.d.\ Gaussian noise
    \STATE Compute rank-$r$ SVD:
           $\mathbf{E}=\mathbf{U}\Sigma\mathbf{V}^\top,\;
       r=\mathrm{rank}(\mathbf{E})$
           \hfill // $\Sigma=\mathrm{diag}(\sigma_1,\dots,\sigma_r)$, $\mathbf{V}\in\mathbb{R}^{d\times r}$
    \STATE $\mathbf{s}\gets\sqrt{\mathrm{diag}(\Sigma)}\in\mathbb{R}^{r}$
           \hfill // singular values
    \STATE $\mu\gets\mathrm{mean}(\mathbf{s})$;
           \quad $\tilde{\mathbf{s}}\gets\mathbf{s}/\mu$
           \hfill // normalize
    \STATE Project noise onto right singular vectors:
            $\mathbf{P} \gets \boldsymbol{\varepsilon} \mathbf{V}\in\mathbb{R}^{B\times r}$
    \STATE Scale by normalized singular values:
            $\boldsymbol{\Delta}_{\text{scaled}} \gets \mathbf{P} \odot \tilde{\mathbf{s}}$
            \hfill  // broadcast $\tilde{\mathbf{s}}$ row-wise
    \STATE Reconstruct perturbation:
           $\Delta\mathbf{E} \gets \boldsymbol{\Delta}_{\text{scaled}} \mathbf{V}^\top\in\mathbb{R}^{B\times d}$
    \STATE Compute per-dimension magnitude:
           $m\gets\alpha/\sqrt{d}$
    \STATE Return augmented embeddings:
           $\mathbf{E}'\gets\mathbf{E}+m\cdot\Delta\mathbf{E}$
  \end{algorithmic}
    \end{algorithm}
  \end{minipage}
\end{figure*}

\begin{table*}[t]
    \centering
        \begin{tabular}{lcccc}
            \toprule
            \textbf{Shared} & \multicolumn{2}{c}{} &  &\\
            \midrule
            Learning rate (LR) & \multicolumn{4}{c}{$2\text{e-5}$}\\
            Warmup Steps& \multicolumn{4}{c}{200}\\
            Training Steps& \multicolumn{4}{c}{2000}\\
            lr Scheduler& \multicolumn{4}{c}{linear}\\
            Normalize& \multicolumn{4}{c}{True}\\
     Precision& \multicolumn{4}{c}{BF16}\\
            Temperature& \multicolumn{4}{c}{0.02}\\
     LoRA Rank& \multicolumn{4}{c}{8}\\
     Noise Strength& \multicolumn{4}{c}{0.1}\\
     Noise Distribution& \multicolumn{4}{c}{Sublinear scaling}\\
            \midrule
            \textbf{Model specific}& Phi3.5-V-4B& Qwen2-VL-2B& LLaVA1.6-LR& LLaVA1.6-HR\\
            \midrule
             Image Resolution& -& high& low&high\\        
             Train Samples& 662K& 1031K& 1031K&1031K\\
             Batch Size Per GPU& 128& 256& 256&256\\
             Num Crops& 4& -& -&-\\
             Max Len& 256& 4096& -& -\\
             Train Hours& 53& 120& 120&236\\
            \bottomrule
        \end{tabular}
    \caption{Training details of FANoise.}
    \label{tab:training hyperparameters}
\end{table*}
\clearpage

\begin{table*}[htbp]
    \resizebox{1\textwidth}{!}{
        \begin{tabular}{lcccccccccc}
            \toprule
             & CLIP&  BLIP2&  UniIR&  MagicLens&  UniME(Phi) &  UniME(LLaVA)  & VLM2Vec(Phi)& FANoise(Phi) &FANoise(Qwen)&VLM2Vec(Qwen)\\
             \toprule
             \multicolumn{11}{c}{Classification(10 tasks)}\\
             \midrule
             ImageNet-1k&  55.8&  10.3&  58.3&  48.0&  66.8&  71.3 & 65.6& 67.8& 79.1& 77.6\\
             N24News&  34.7&  36.0&  42.5&  33.7&  79.2&  79.5 & 79.5& 81.0& 74.7& 76.3\\
             HatefulMems&  51.1&  49.6&  56.4&  49.0&  65.4&  64.6 & 67.1& 68.4& 59.7& 62.1\\
             VOC2007&  50.7&  52.1&  66.2&  51.6&  90.6&  90.4 & 88.6& 90.3& 81.0& 79.9\\
             SUN397&  43.4&  34.5&  63.2&  57.0&  72.7&  75.9  & 72.7& 72.8& 75.9& 74.3\\
             Place365&  28.5&  21.5&  36.5&  31.5&  41.9&  45.6 & 42.6& 41.2& 34.9& 36.1\\
             ImageNet-A&  25.5&  3.2&  9.8&  8.0&  17.9&  45.5 & 19.3& 17.4& 56.0& 51.5\\
             ImageNet-R&  75.6&  39.7&  66.2&  70.9&  72.5&  78.4 & 70.2& 69.4& 84.1& 86.6\\
             ObjectNet&  43.4&  20.6&  32.2&  31.6&  26.7&  36.4 & 29.5& 25& 29.8& 23.5\\
             Country-211&  19.2&  2.5&  11.3&  6.2&  12.2&  18.7  & 13.0& 12.3& 27.8& 22.0\\
             All Classification&  42.8&  27.0&  44.3&  38.8&  54.6&  60.6  & 54.8& 54.6& 60.3& 59.0\\
             \midrule
             \multicolumn{11}{c}{VQA (10 tasks)}\\
             \midrule
             OK-VQA&  7.5&  8.7&  25.4&  12.7&  64.6&  68.3  & 63.2& 64.3& 50.9& 47.4\\
             A-OKVQA&  3.8&  3.2&  8.8&  2.9&  51.6&  58.7  & 50.2& 51.9& 44.9& 40.5\\
             DocVQA&  4.0&  2.6&  6.2&  3.0&  81.6&  67.6  & 78.4& 80.5& 88.0& 85.2\\
             InfographicsVQA&  4.6&  2.0&  4.6&  5.9&  42.1&  37.0  & 40.8& 43.6& 52.6& 49.7\\
             ChartQA&  1.4&  0.5&  1.6&  0.9&  58.5&  33.4  & 59.0& 59.1& 44.8& 42.5\\
             Visual7W&  4.0&  1.3&  14.5&  2.5&  51.1&  51.7  & 47.7& 52.6& 49.4& 50.3\\
             ScienceQA&  9.4&  6.8&  12.8&  5.2&  42.4&  40.5  & 43.4& 44.2& 30.8& 29.6\\
             VizWiz&  8.2&  4.0&  24.3&  1.7&  38.9&  42.7  & 39.2& 41.1& 38.7& 37.0\\
             GQA&  41.3&  9.7&  48.8&  43.5&  61.0&  63.6  & 60.7& 56.4& 45.3& 48.0\\
             TextVQA&  7.0&  3.3&  15.1&  4.6&  67.1&  65.2  & 66.1& 67.6& 67.7& 63.3\\
             All VQA&  9.1&  4.2&  16.2&  8.3&  55.9&  52.9  & 54.9& 56.1& 51.3& 49.4\\
             \midrule
             \multicolumn{11}{c}{Retrieval (12 tasks)}\\
             \midrule
             VisDial&  30.7&  18.0&  42.2&  24.8&  76.3&  79.7  & 73.3& 76.5& 78.2& 76.1\\
             CIRR&  12.6&  9.8&  51.3&  39.1&  47.9&  52.2  & 47.8& 53& 50.6& 48.6\\
             VisualNews\_t2i&  78.9&  48.1&  74.3&  50.7&  68.1&  74.8  & 67.2& 68.7& 74.1& 74.8\\
             VisualNews\_i2t&  79.6&  13.5&  76.8&  21.1&  72.2&  78.8  & 70.7& 70.8& 76.6& 74.7\\
             MSCOCO\_t2i&  59.5&  53.7&  68.5&  54.1&  71.6&  74.9  & 70.6& 72.6& 73.3& 71.5\\
             MSCOCO\_i2t&  57.7&  20.3&  72.1&  40.0&  69.0&  73.8  & 66.5& 68.7& 70.7& 68.6\\
             NIGHTS&  60.4&  56.5&  66.2&  58.1&  69.1&  66.2  & 66.1& 68.6& 66.1& 65.9\\
             WebQA&  67.5&  55.4&  89.6&  43.0&  90.0&  89.8  & 88.1& 87.8& 86.8& 86.8\\
             FahsionIQ&  11.4&  9.3&  40.2&  11.2&  15.6&  16.5  & 12.9& 11.9& 17.7& 13.9\\
             Wiki-SS-NQ&  55.0&  28.7&  12.2&  18.7&  60.5&  66.6  & 56.6& 59.0& 61.0& 57.7\\
             OVEN&  41.1&  39.5&  69.4&  1.6&  49.0&  55.7  & 47.3& 49.1& 67.0& 66.3\\
             EDIS&  81.0&  54.4&  79.2&  62.6&  84.5&  86.2  & 79.9& 80.3& 71.7& 80.3\\
             All Retrieval&  53.0&  33.9&  61.8&  35.4&  64.5&  67.9  & 62.3& 63.9& 66.2& 65.4\\
             \midrule
             \multicolumn{11}{c}{Visual Grounding (4 tasks)}\\
             \midrule
             MSCOCO&  33.8&  28.9&  46.6&  22.1&  74.8&  76.5  & 67.3& 65.0& 66.5& 67.5\\
             RefCOCO&  56.9&  47.4&  67.8&  22.8&  86.9&  89.3  & 84.7& 83.8& 78.4& 80.9\\
             RefCOCO-matching&  61.3&  59.5&  62.9&  35.6&  81.7&  90.6  & 79.2& 78.6& 74.6& 75.6\\
             Visual7W-pointing&  55.1&  52.0&  71.3&  23.4&  83.8&  84.1  & 86.8& 86.8& 69.6& 69.7\\
             All Visual Grounding&  51.8&  47.0&  65.3&  26.0&  81.8&  85.1  & 79.5& 78.6& 72.3& 73.4\\
             \midrule
             \multicolumn{11}{c}{Final Score (36 tasks)}\\
             \midrule
             All&  37.8&  25.2&  44.7&  27.8&  61.3&   63.6 & 60.1& 60.8& 61.1& 60.1\\
             All IND&  37.1&  25.3&  47.1&  31.0&  68.2&  68.4  & 66.5& 68.2& 67.2& 66.0\\
             All OOD& 38.7& 25.1& 41.7& 23.7& 52.7& 57.9 & 52.0& 51.5& 53.4& 52.6\\
             \bottomrule
        \end{tabular}
    }
    \caption{The detailed results of the baselines and FANoise on MMEB. The out-of-distribution datasets are highlighted with yellow background in the table."Qwen" refers to Qwen2-VL-2B-Instruct, "LLaVA" uses the LLaVA-1.6 model, "Phi" refers to Phi3.5-V.}
    \label{tab:all mmeb result}
\end{table*}
\clearpage

\bibliography{arxiv}        

@article{liu2022universal,
  title={Universal vision-language dense retrieval: Learning a unified representation space for multi-modal retrieval},
  author={Liu, Zhenghao and Xiong, Chenyan and Lv, Yuanhuiyi and Liu, Zhiyuan and Yu, Ge},
  journal={arXiv preprint arXiv:2209.00179},
  year={2022}
}

@inproceedings{wei2024uniir,
  title={Uniir: Training and benchmarking universal multimodal information retrievers},
  author={Wei, Cong and Chen, Yang and Chen, Haonan and Hu, Hexiang and Zhang, Ge and Fu, Jie and Ritter, Alan and Chen, Wenhu},
  booktitle={European Conference on Computer Vision},
  pages={387--404},
  year={2024},
  organization={Springer}
}

@article{wang2024qwen2,
  title={Qwen2-vl: Enhancing vision-language model's perception of the world at any resolution},
  author={Wang, Peng and Bai, Shuai and Tan, Sinan and Wang, Shijie and Fan, Zhihao and Bai, Jinze and Chen, Keqin and Liu, Xuejing and Wang, Jialin and Ge, Wenbin and others},
  journal={arXiv preprint arXiv:2409.12191},
  year={2024}
}

@misc{liu2024llavanext,
  title={Llavanext: Improved reasoning, ocr, and world knowledge},
  author={Liu, Haotian and Li, Chunyuan and Li, Yuheng and Li, Bo and Zhang, Yuanhan and Shen, Sheng and Lee, Yong Jae},
  year={2024}
}

@article{jiang2024vlm2vec,
  title={Vlm2vec: Training vision-language models for massive multimodal embedding tasks},
  author={Jiang, Ziyan and Meng, Rui and Yang, Xinyi and Yavuz, Semih and Zhou, Yingbo and Chen, Wenhu},
  journal={arXiv preprint arXiv:2410.05160},
  year={2024}
}

@article{zhang2024gme,
  title={GME: Improving Universal Multimodal Retrieval by Multimodal LLMs},
  author={Zhang, Xin and Zhang, Yanzhao and Xie, Wen and Li, Mingxin and Dai, Ziqi and Long, Dingkun and Xie, Pengjun and Zhang, Meishan and Li, Wenjie and Zhang, Min},
  journal={arXiv preprint arXiv:2412.16855},
  year={2024}
}

@article{chen2025mme5,
  title={mmE5: Improving Multimodal Multilingual Embeddings via High-quality Synthetic Data},
  author={Chen, Haonan and Wang, Liang and Yang, Nan and Zhu, Yutao and Zhao, Ziliang and Wei, Furu and Dou, Zhicheng},
  journal={arXiv preprint arXiv:2502.08468},
  year={2025}
}

@misc{jain2023neftune,
  title={Neftune: Noisy Embeddings Improve Instruction Finetuning. arXiv, abs/2310.05914},
  author={Jain, N and Chiang, P-y and Wen, Y and Kirchenbauer, J and Chu, HM and Somepalli, G and Bartoldson, BR and Kailkhura, B and Schwarzschild, A and Saha, A},
  year={2023}
}

@inproceedings{gao2021simcse,
   title={{SimCSE}: Simple Contrastive Learning of Sentence Embeddings},
   author={Gao, Tianyu and Yao, Xingcheng and Chen, Danqi},
   booktitle={Empirical Methods in Natural Language Processing (EMNLP)},
   year={2021}
}

@inproceedings{radford2021clip,
  title={Learning transferable visual models from natural language supervision},
  author={Radford, Alec and Kim, Jong Wook and Hallacy, Chris and Ramesh, Aditya and Goh, Gabriel and Agarwal, Sandhini and Sastry, Girish and Askell, Amanda and Mishkin, Pamela and Clark, Jack and others},
  booktitle={International conference on machine learning},
  pages={8748--8763},
  year={2021},
  organization={PmLR}
}

@inproceedings{li2022blip,
  title={Blip: Bootstrapping language-image pre-training for unified vision-language understanding and generation},
  author={Li, Junnan and Li, Dongxu and Xiong, Caiming and Hoi, Steven},
  booktitle={International conference on machine learning},
  pages={12888--12900},
  year={2022},
  organization={PMLR}
}

@inproceedings{li2023blip,
  title={Blip-2: Bootstrapping language-image pre-training with frozen image encoders and large language models},
  author={Li, Junnan and Li, Dongxu and Savarese, Silvio and Hoi, Steven},
  booktitle={International conference on machine learning},
  pages={19730--19742},
  year={2023},
  organization={PMLR}
}

@inproceedings{jia2021align,
  title={Scaling up visual and vision-language representation learning with noisy text supervision},
  author={Jia, Chao and Yang, Yinfei and Xia, Ye and Chen, Yi-Ting and Parekh, Zarana and Pham, Hieu and Le, Quoc and Sung, Yun-Hsuan and Li, Zhen and Duerig, Tom},
  booktitle={International conference on machine learning},
  pages={4904--4916},
  year={2021},
  organization={PMLR}
}

@inproceedings{zhai2023sigmoid,
  title={Sigmoid loss for language image pre-training},
  author={Zhai, Xiaohua and Mustafa, Basil and Kolesnikov, Alexander and Beyer, Lucas},
  booktitle={Proceedings of the IEEE/CVF international conference on computer vision},
  pages={11975--11986},
  year={2023}
}

@article{ren2024vista,
  title={VISTA: Enhancing Long-Duration and High-Resolution Video Understanding by Video Spatiotemporal Augmentation},
  author={Ren, Weiming and Yang, Huan and Min, Jie and Wei, Cong and Chen, Wenhu},
  journal={arXiv preprint arXiv:2412.00927},
  year={2024}
}

@article{zhang2024magiclens,
  title={Magiclens: Self-supervised image retrieval with open-ended instructions},
  author={Zhang, Kai and Luan, Yi and Hu, Hexiang and Lee, Kenton and Qiao, Siyuan and Chen, Wenhu and Su, Yu and Chang, Ming-Wei},
  journal={arXiv preprint arXiv:2403.19651},
  year={2024}
}

@article{zhou2024megapairs,
  title={MegaPairs: Massive Data Synthesis For Universal Multimodal Retrieval},
  author={Zhou, Junjie and Liu, Zheng and Liu, Ze and Xiao, Shitao and Wang, Yueze and Zhao, Bo and Zhang, Chen Jason and Lian, Defu and Xiong, Yongping},
  journal={arXiv preprint arXiv:2412.14475},
  year={2024}
}

@article{jiang2024e5v,
  title={E5-v: Universal embeddings with multimodal large language models},
  author={Jiang, Ting and Song, Minghui and Zhang, Zihan and Huang, Haizhen and Deng, Weiwei and Sun, Feng and Zhang, Qi and Wang, Deqing and Zhuang, Fuzhen},
  journal={arXiv preprint arXiv:2407.12580},
  year={2024}
}

@article{oord2018infonce,
  title={Representation learning with contrastive predictive coding},
  author={Oord, Aaron van den and Li, Yazhe and Vinyals, Oriol},
  journal={arXiv preprint arXiv:1807.03748},
  year={2018}
}

@article{gao2021GradCache,
  title={Scaling deep contrastive learning batch size under memory limited setup},
  author={Gao, Luyu and Zhang, Yunyi and Han, Jiawei and Callan, Jamie},
  journal={arXiv preprint arXiv:2101.06983},
  year={2021}
}

@article{hu2022lora,
  title={Lora: Low-rank adaptation of large language models.},
  author={Hu, Edward J and Shen, Yelong and Wallis, Phillip and Allen-Zhu, Zeyuan and Li, Yuanzhi and Wang, Shean and Wang, Lu and Chen, Weizhu and others},
  journal={ICLR},
  volume={1},
  number={2},
  pages={3},
  year={2022}
}

@misc{bge_embedding,
      title={C-Pack: Packaged Resources To Advance General Chinese Embedding}, 
      author={Shitao Xiao and Zheng Liu and Peitian Zhang and Niklas Muennighoff},
      year={2023},
      eprint={2309.07597},
      archivePrefix={arXiv},
      primaryClass={cs.CL}
}

@article{li2023gte,
  title={Towards general text embeddings with multi-stage contrastive learning},
  author={Li, Zehan and Zhang, Xin and Zhang, Yanzhao and Long, Dingkun and Xie, Pengjun and Zhang, Meishan},
  journal={arXiv preprint arXiv:2308.03281},
  year={2023}
}

@article{lan2025llave,
  title={LLaVE: Large Language and Vision Embedding Models with Hardness-Weighted Contrastive Learning},
  author={Lan, Zhibin and Niu, Liqiang and Meng, Fandong and Zhou, Jie and Su, Jinsong},
  journal={arXiv preprint arXiv:2503.04812},
  year={2025}
}

@article{gu2025breaking,
  title={Breaking the Modality Barrier: Universal Embedding Learning with Multimodal LLMs},
  author={Gu, Tiancheng and Yang, Kaicheng and Feng, Ziyong and Wang, Xingjun and Zhang, Yanzhao and Long, Dingkun and Chen, Yingda and Cai, Weidong and Deng, Jiankang},
  journal={arXiv preprint arXiv:2504.17432},
  year={2025}
}

@inproceedings{2021zhangCELoss,
  title={Learning with Different Amounts of Annotation: From Zero to Many Labels},
  author={ Zhang, Shujian  and  Gong, Chengyue  and  Choi, Eunsol },
  booktitle={Empirical Methods in Natural Language Processing},
  year={2021},
}

@inproceedings{wang2019symmetric,
  title={Symmetric cross entropy for robust learning with noisy labels},
  author={Wang, Yisen and Ma, Xingjun and Chen, Zaiyi and Luo, Yuan and Yi, Jinfeng and Bailey, James},
  booktitle={Proceedings of the IEEE/CVF international conference on computer vision},
  pages={322--330},
  year={2019}
}

@article{2023Dynamics,
  title={Dynamics-aware loss for learning with label noise},
  author={ Li, Xiu Chuan  and  Xia, Xiaobo  and  Zhu, Fei  and  Liu, Tongliang  and  Zhang, Xu Yao  and  Liu, Cheng Lin },
  journal={Pattern Recognition},
  volume={144},
  number={000},
  pages={11},
  year={2023},
}

@article{shorten2019survey,
  title={A survey on image data augmentation for deep learning},
  author={Shorten, Connor and Khoshgoftaar, Taghi M},
  journal={Journal of big data},
  volume={6},
  number={1},
  pages={1--48},
  year={2019},
  publisher={Springer}
}

@inproceedings{szegedy2016rethinking,
  title={Rethinking the inception architecture for computer vision},
  author={Szegedy, Christian and Vanhoucke, Vincent and Ioffe, Sergey and Shlens, Jon and Wojna, Zbigniew},
  booktitle={Proceedings of the IEEE conference on computer vision and pattern recognition},
  pages={2818--2826},
  year={2016}
}

@article{liu2020early,
  title={Early-learning regularization prevents memorization of noisy labels},
  author={Liu, Sheng and Niles-Weed, Jonathan and Razavian, Narges and Fernandez-Granda, Carlos},
  journal={Advances in neural information processing systems},
  volume={33},
  pages={20331--20342},
  year={2020}
}

@inproceedings{kim2019nlnl,
  title={Nlnl: Negative learning for noisy labels},
  author={Kim, Youngdong and Yim, Junho and Yun, Juseung and Kim, Junmo},
  booktitle={Proceedings of the IEEE/CVF international conference on computer vision},
  pages={101--110},
  year={2019}
}

@article{han2018co,
  title={Co-teaching: Robust training of deep neural networks with extremely noisy labels},
  author={Han, Bo and Yao, Quanming and Yu, Xingrui and Niu, Gang and Xu, Miao and Hu, Weihua and Tsang, Ivor and Sugiyama, Masashi},
  journal={Advances in neural information processing systems},
  volume={31},
  year={2018}
}

@inproceedings{xia2023combating,
  title={Combating noisy labels with sample selection by mining high-discrepancy examples},
  author={Xia, Xiaobo and Han, Bo and Zhan, Yibing and Yu, Jun and Gong, Mingming and Gong, Chen and Liu, Tongliang},
  booktitle={Proceedings of the IEEE/CVF international conference on computer vision},
  pages={1833--1843},
  year={2023}
}

@article{lu2022selc,
  title={SELC: self-ensemble label correction improves learning with noisy labels},
  author={Lu, Yangdi and He, Wenbo},
  journal={arXiv preprint arXiv:2205.01156},
  year={2022}
}

@article{gong2022class,
  title={Class-wise denoising for robust learning under label noise},
  author={Gong, Chen and Ding, Yongliang and Han, Bo and Niu, Gang and Yang, Jian and You, Jane and Tao, Dacheng and Sugiyama, Masashi},
  journal={IEEE Transactions on Pattern Analysis and Machine Intelligence},
  volume={45},
  number={3},
  pages={2835--2848},
  year={2022},
  publisher={IEEE}
}

@article{zhang2021learning,
  title={Learning with feature-dependent label noise: A progressive approach},
  author={Zhang, Yikai and Zheng, Songzhu and Wu, Pengxiang and Goswami, Mayank and Chen, Chao},
  journal={arXiv preprint arXiv:2103.07756},
  year={2021}
}

@inproceedings{chen2020simclr,
  title={A simple framework for contrastive learning of visual representations},
  author={Chen, Ting and Kornblith, Simon and Norouzi, Mohammad and Hinton, Geoffrey},
  booktitle={International conference on machine learning},
  pages={1597--1607},
  year={2020},
  organization={PmLR}
}

@inproceedings{he2020momentum,
  title={Momentum contrast for unsupervised visual representation learning},
  author={He, Kaiming and Fan, Haoqi and Wu, Yuxin and Xie, Saining and Girshick, Ross},
  booktitle={Proceedings of the IEEE/CVF conference on computer vision and pattern recognition},
  pages={9729--9738},
  year={2020}
}

@article{ho2020contrastive,
  title={Contrastive learning with adversarial examples},
  author={Ho, Chih-Hui and Nvasconcelos, Nuno},
  journal={Advances in Neural Information Processing Systems},
  volume={33},
  pages={17081--17093},
  year={2020}
}

@article{izmailov2018swa,
  title={Averaging weights leads to wider optima and better generalization},
  author={Izmailov, Pavel and Podoprikhin, Dmitrii and Garipov, Timur and Vetrov, Dmitry and Wilson, Andrew Gordon},
  journal={arXiv preprint arXiv:1803.05407},
  year={2018}
}

@inproceedings{verma2019manifold,
  title={Manifold mixup: Better representations by interpolating hidden states},
  author={Verma, Vikas and Lamb, Alex and Beckham, Christopher and Najafi, Amir and Mitliagkas, Ioannis and Lopez-Paz, David and Bengio, Yoshua},
  booktitle={International conference on machine learning},
  pages={6438--6447},
  year={2019},
  organization={PMLR}
}

@article{grill2020bootstrap,
  title={Bootstrap your own latent-a new approach to self-supervised learning},
  author={Grill, Jean-Bastien and Strub, Florian and Altch{\'e}, Florent and Tallec, Corentin and Richemond, Pierre and Buchatskaya, Elena and Doersch, Carl and Avila Pires, Bernardo and Guo, Zhaohan and Gheshlaghi Azar, Mohammad and others},
  journal={Advances in neural information processing systems},
  volume={33},
  pages={21271--21284},
  year={2020}
}

@article{ho2020denoising,
  title={Denoising diffusion probabilistic models},
  author={Ho, Jonathan and Jain, Ajay and Abbeel, Pieter},
  journal={Advances in neural information processing systems},
  volume={33},
  pages={6840--6851},
  year={2020}
}

@inproceedings{vincent2008extracting,
  title={Extracting and composing robust features with denoising autoencoders},
  author={Vincent, Pascal and Larochelle, Hugo and Bengio, Yoshua and Manzagol, Pierre-Antoine},
  booktitle={Proceedings of the 25th international conference on Machine learning},
  pages={1096--1103},
  year={2008}
}

@inproceedings{wu2018unsupervised,
  title={Unsupervised feature learning via non-parametric instance discrimination},
  author={Wu, Zhirong and Xiong, Yuanjun and Yu, Stella X and Lin, Dahua},
  booktitle={Proceedings of the IEEE conference on computer vision and pattern recognition},
  pages={3733--3742},
  year={2018}
}

@article{oord2018representation,
  title={Representation learning with contrastive predictive coding},
  author={Oord, Aaron van den and Li, Yazhe and Vinyals, Oriol},
  journal={arXiv preprint arXiv:1807.03748},
  year={2018}
}

@article{caron2020unsupervised,
  title={Unsupervised learning of visual features by contrasting cluster assignments},
  author={Caron, Mathilde and Misra, Ishan and Mairal, Julien and Goyal, Priya and Bojanowski, Piotr and Joulin, Armand},
  journal={Advances in neural information processing systems},
  volume={33},
  pages={9912--9924},
  year={2020}
}

@article{bao2021beit,
  title={Beit: Bert pre-training of image transformers},
  author={Bao, Hangbo and Dong, Li and Piao, Songhao and Wei, Furu},
  journal={arXiv preprint arXiv:2106.08254},
  year={2021}
}

@article{logeswaran2018efficient,
  title={An efficient framework for learning sentence representations},
  author={Logeswaran, Lajanugen and Lee, Honglak},
  journal={arXiv preprint arXiv:1803.02893},
  year={2018}
}

@article{velivckovic2018deep,
  title={Deep graph infomax},
  author={Veli{\v{c}}kovi{\'c}, Petar and Fedus, William and Hamilton, William L and Li{\`o}, Pietro and Bengio, Yoshua and Hjelm, R Devon},
  journal={arXiv preprint arXiv:1809.10341},
  year={2018}
}

@article{abdin2024phi,
  title={Phi-3 technical report: A highly capable language model locally on your phone},
  author={Abdin, Marah and Aneja, Jyoti and Awadalla, Hany and Awadallah, Ahmed and Awan, Ammar Ahmad and Bach, Nguyen and Bahree, Amit and Bakhtiari, Arash and Bao, Jianmin and Behl, Harkirat and others},
  journal={arXiv preprint arXiv:2404.14219},
  year={2024}
}

@book{tao2012topics,
  title={Topics in random matrix theory},
  author={Tao, Terence},
  volume={132},
  year={2012},
  publisher={American Mathematical Soc.}
}

@article{perry2016optimal,
  title={Optimality and sub-optimality of PCA for spiked random matrices and synchronization},
  author={Perry, Amelia and Wein, Alexander S and Bandeira, Afonso S and Moitra, Ankur},
  journal={arXiv preprint arXiv:1609.05573},
  year={2016}
}

@article{perry2018optimality,
  title={Optimality and sub-optimality of PCA I: Spiked random matrix models},
  author={Perry, Amelia and Wein, Alexander S and Bandeira, Afonso S and Moitra, Ankur},
  journal={The Annals of Statistics},
  volume={46},
  number={5},
  pages={2416--2451},
  year={2018},
  publisher={JSTOR}
}

@article{marchenko1967distribution,
  title={Distribution of eigenvalues for some sets of random matrices},
  author={Marchenko, Vladimir A and Pastur, Leonid A},
  journal={Mathematics of the USSR-Sbornik},
  year={1967}
}

@article{li2024onevision,
  title={Llava-onevision: Easy visual task transfer},
  author={Li, Bo and Zhang, Yuanhan and Guo, Dong and Zhang, Renrui and Li, Feng and Zhang, Hao and Zhang, Kaichen and Zhang, Peiyuan and Li, Yanwei and Liu, Ziwei and others},
  journal={arXiv preprint arXiv:2408.03326},
  year={2024}
}


\end{document}